\documentclass[runningheads]{llncs}

\usepackage[mobile]{eccv}

\usepackage{eccvabbrv}
\usepackage[dvipsnames]{xcolor}
\usepackage{graphicx}
\usepackage{booktabs}
\usepackage{makecell}

\usepackage{amssymb}
\usepackage{pifont}
\newcommand{\cmark}{\ding{51}}%
\newcommand{\xmark}{\ding{55}}%

\usepackage{tikz}
\usepackage{enumitem}
\usepackage{tikzpagenodes}

\usepackage[breaklinks,colorlinks,citecolor=eccvblue]{hyperref}

\newcommand{\method}{\texttt{BRAVE}\xspace}

\newcommand{\bridge}{\texttt{MEQ-Former}\xspace}

\newcommand{\weburl}{\url{https://brave-vlms.epfl.ch}\xspace}

\begin{document}

\title{BRAVE~~~~: Broadening the visual encoding of vision-language models}



\authorrunning{O.F.~Kar et al.}
\titlerunning{BRAVE: Broadening the visual encoding of vision-language models}


\author{
\quad {O\u{g}uzhan Fatih Kar}$^{1,2}$ \quad  {Alessio Tonioni}$^{1}$  \quad  {Petra Poklukar}$^{1}$  \vspace{0.3em} \\
{Achin Kulshrestha}$^{1}$ \quad  {Amir Zamir}$^{2}$  \quad  {Federico Tombari}$^{1}$ 
}

\institute{$^1$Google \quad $^2$Swiss Federal Institute of Technology Lausanne~(EPFL) \vspace{0.5em} \\ \weburl}

\maketitle

\vspace{-0.2in}
\begin{figure}[h!]
  \centering
  \includegraphics[width=\textwidth]{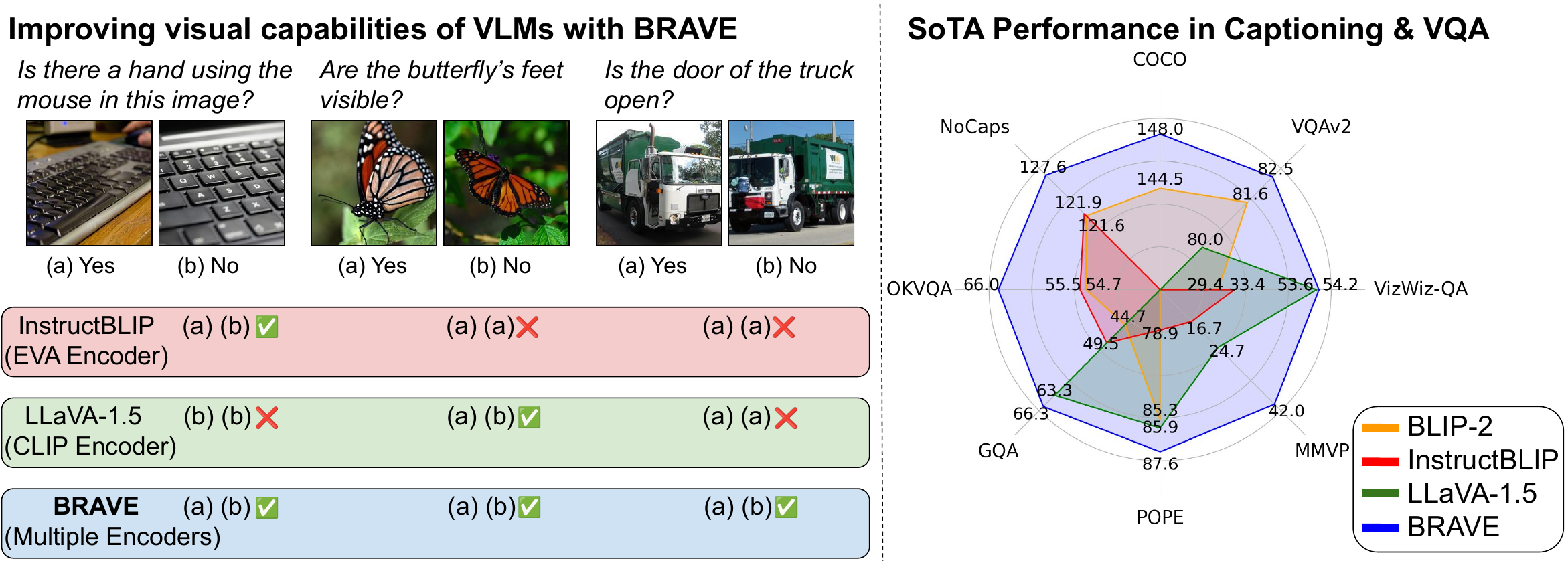}
  \caption{
  We propose \method to broaden the visual capabilities of vision-language models~(VLMs). \textbf{Left:} In contrast to existing methods, e.g. InstructBLIP~\cite{dai2023instructblip} or LLaVA-1.5~\cite{liu2023improved}, that use a single vision encoder~\cite{li2023evaluating, tong2024eyes}, \method combines diverse features from multiple vision encoders into a more versatile and compact representation. The examples are taken from~\cite{tong2024eyes} and assess the VLM's ability to differentiate images with visual differences. \textbf{Right:} \method leads to state-of-the-art performance on a wide range of captioning and visual question answering tasks. Furthermore, it significantly improves the performance on benchmarks, e.g. MMVP, where commonly employed vision encoders, e.g. CLIP, fail.
  }\vspace{-0.6in}
  \label{fig:pull}
\end{figure}

\begin{tikzpicture}[remember picture,overlay,shift={(current page.north west)}]
\node[anchor=north west,xshift=3.5cm,yshift=-.9cm]{\includegraphics[width=0.7cm]{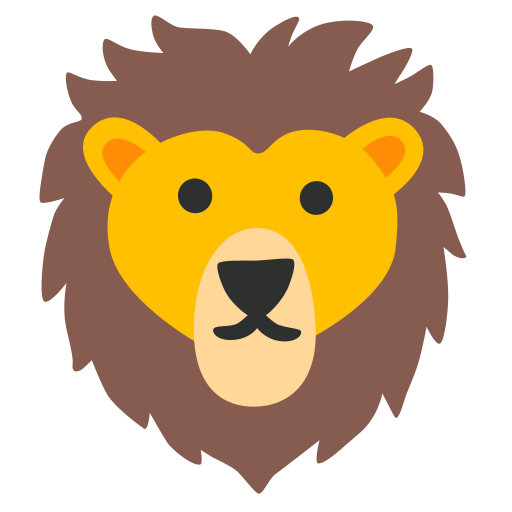}};
\end{tikzpicture}

\begin{abstract}

Vision-language models (VLMs) are typically composed of a vision encoder, e.g. CLIP, and a language model~(LM) that interprets the encoded features to solve downstream tasks. Despite remarkable progress, VLMs are subject to several shortcomings due to the limited capabilities of vision encoders, e.g. ``blindness'' to certain image features, visual hallucination, etc. To address these issues, we study broadening the visual encoding capabilities of VLMs. We first comprehensively benchmark several vision encoders with different inductive biases for solving VLM tasks. We observe that there is no single encoding configuration that consistently achieves top performance across different tasks, and encoders with different biases can perform surprisingly similarly. Motivated by this, we introduce a method, named \method, that consolidates features from multiple frozen encoders into a more versatile representation that can be directly fed as the input to a frozen LM. \method achieves state-of-the-art performance on a broad range of captioning and VQA benchmarks and significantly reduces the aforementioned issues of VLMs, while requiring a smaller number of trainable parameters than existing methods and having a more compressed representation. Our results highlight the potential of incorporating different visual biases for a more broad and contextualized visual understanding of VLMs.

\end{abstract}

\vspace{-0.4in}
\section{Introduction}
\label{sec:intro}

Vision-language models~(VLMs) have recently seen significant improvements on solving tasks requiring both visual and text understanding capabilities such as captioning, visual question answering~(VQA), and instruction following. These advancements are fueled by the progress in uni-modal vision encoders~\cite{radford2021clip,fang2023eva} and language models~(LMs)~\cite{chung2022scaling, chiang2023vicuna}, which are then combined with different ``bridging'' techniques~\cite{Alayrac2022Flamingo,liu2023visual,li2023blip2} to create VLMs. Improved training recipes~\cite{dai2023instructblip, liu2023improved} and scalability efforts in terms of training data and model size~\cite{chen2022pali, zhai2022scaling, wang2022git, Schuhmann2022LAION5B, bai2023qwen} further boost the performance. 

Despite the progress, VLMs are subject to several shortcomings: on the language side, the LMs are known to be susceptible to hallucinations and logical faults~\cite{bang2023multitask, shen2023chatgpt,thorp2023chatgpt,guo2023close}, while on the vision side, they are limited by the capabilities of the vision encoder. For example, Tong et al.\cite{tong2024eyes} observed that commonly employed vision encoders such as CLIP~\cite{radford2021clip} can exhibit ``blindness'', i.e. fail to differentiate images with clear visual differences~(See Fig.~\ref{fig:pull}, left). Similarly, Li et al.\cite{li2023evaluating} showed that VLMs are prone to visual hallucinations where the models imagine incorrect details about a given image. These limitations create a bottleneck for developing performant and visually grounded VLMs.

To address these, we study broadening the visual encoding capabilities of VLMs. It is known in machine learning that different representations of an input can yield different generalization properties, and each can play a role when ensembled to create a more complete representation~\cite{geman1992neural}. Inspired by this, we first perform a comprehensive evaluation of VLMs which differ only in their vision encoders, i.e. \textit{visual biases}. For this, we consider encoders with different biases, e.g. different objectives, training data, and model sizes. Our results in Table~\ref{tab:benchmark} show that
different encoders lead to varying performance across vision-language tasks and there is no single encoder achieving a consistent top performance across tasks.
Furthermore, we find that encoders with different biases can perform surprisingly similarly, suggesting that there could be multiple cues~\cite{goodwin1995seeing,howard1995binocular} that can be leveraged to solve the vision-language tasks.

Motivated by these findings, we propose to employ various vision encoders for VLMs and introduce a method to learn how to combine them \textit{efficiently}.
We denote the method as \method, which stands for \textit{\textbf{br}o\textbf{a}dening the \textbf{v}isual \textbf{e}ncoding of VLMs}.
As shown in Fig.~\ref{fig:method}, \method combines features from an \textit{arbitrary} number of vision encoders into a compressed fixed-length visual representation which is then fed as a soft visual prompt to a frozen LM. We achieve the bridging of encoders and the LM by introducing a lightweight \textit{\textbf{m}ulti-\textbf{e}ncoder \textbf{q}uerying trans\textbf{former}}~(\bridge), 
which is a multi-encoder generalization of the Q-Former proposed in the single encoder BLIP-2 framework~\cite{li2023blip2}.
In particular, \bridge accepts a textual prompt and a set of learnable queries as inputs and jointly cross-attends to the features from different vision encoders~(Fig.~\ref{fig:method}). We provide a simple yet effective recipe in Sec.~\ref{sec:method} for training the \bridge in a single stage that bypasses the two-stage pre-training paradigm of~\cite{li2023blip2}. Notably, this is achieved with a smaller number of trainable parameters during pre-training than the existing methods~\cite{liu2023improved, dai2023instructblip, lin2023sphinx, bai2023qwen, chen2022pali, Alayrac2022Flamingo, wang2022git}~(Sec.~\ref{sec:results}). 

\looseness=-1
Our extensive evaluation on a wide range of captioning and VQA tasks shows that \method effectively consolidates diverse visual signals into a broad and contextual representation, leading to consistently better performance over the state-of-the-art and improved robustness against out-of-distribution inputs, as shown in Fig.~\ref{fig:pull} and Sec.~\ref{sec:results}. Furthermore, in contrast to previous works that benefit from scaling the language axis~\cite{Alayrac2022Flamingo, chen2022pali,  dai2023instructblip, liu2023improved} by using larger LMs, our work demonstrates that scaling along the vision axis also has a significant potential for VLMs. 

Our contributions can be summarized as follows:

\begin{itemize}
    \item We conduct a systematic analysis of several vision encoders with different inductive biases, e.g. training data, objective, model size, on solving vision-language tasks under a unified training and evaluation framework.
    \item We introduce \method, a method that efficiently consolidates features from any number of vision encoders into a compressed and contextual representation, leading to \textbf{1)} state-of-the-art performance on several captioning and VQA tasks and \textbf{2)} significantly improved robustness against visual hallucinations and out-of-distribution inputs.
    \item We perform a comprehensive ablation study of \method, highlighting the impact of the design choices, which we hope provide useful insights to the researchers for further advancements in VLMs.
\end{itemize}

\section{Impact of vision encoders for vision-language models}\label{sec:benchmark}

To quantify the impact of visual biases on the performance of VLMs, we compare VLMs with different vision encoders on commonly evaluated VQA and captioning tasks. For this, we develop a pre-training, fine-tuning and evaluation setup, as explained next. To the best of our knowledge, this is the first unified and comprehensive evaluation of different vision encoders for vision-language understanding and generation tasks.

\subsubsection{VLM architecture.} We use a frozen vision encoder which is connected to a frozen language model by a bridge network with trainable parameters. For the bridge, we adopt the Q-Former module from~\cite{li2023blip2}. This choice is based on the following reasons: \textbf{1)} the Q-Former resamples visual features to a fixed-length output before feeding them to the LM which enables a fair comparison of vision encoders with different output dimensionalities, \textbf{2)} the Q-Former is efficient to train and evaluate as it is compatible with frozen pre-trained vision encoders and LMs. 
For the vision encoder, we consider a diverse set of options as described in the next paragraph. For the LM, we follow~\cite{li2023blip2} and use FlanT5-XL~\cite{chung2022scaling}. The Q-Former has 110M parameters consisting of a series of cross- and self-attention layers and a fixed number of learnable queries. Unless specified otherwise, we use 32 queries of dimension $768$ which is also the hidden dimension of the Q-Former. The queries interact with visual features through cross-attention layers, enabling a significant reduction in their dimensionality.
For example, CLIP-L/14~\cite{radford2021clip} visual features are reduced from $257\times 1024$ to $32 \times 768$. The output of the Q-Former is then linearly projected to the textual embedding space of the LM to create a \textit{soft visual prompt}, and then prepended to the textual prompt that specifies the task. Both visual and textual prompts are fed as inputs to the frozen LM. For captioning, we use the textual prompt ``\textit{A photo of }'' and for VQA, we directly use the question as the prompt. Similar to InstructBLIP~\cite{dai2023instructblip}, we give the textual prompt as additional input to the Q-Former together with the learnable queries to extract more task-aligned visual prompts.

\looseness=-1
\subsubsection{Vision encoders.} As summarized in Table~\ref{tab:encoder_stats}, we consider eight recently introduced vision encoders: CLIP~\cite{radford2021clip}, OpenCLIP~\cite{cherti2023reproducible}, EVA-CLIP~\cite{fang2023eva}, SIGLIP~\cite{zhai2023sigmoid}, SILC~\cite{naeem2023silc}, ViT-e~\cite{chen2022pali}, ViT-G~\cite{zhai2022scaling}, and DINOv2~\cite{oquab2023dinov2}. While they all use ViT-based backbones~\cite{Dosovitskiy2020vit}, they differ in terms of \textbf{1)} training data, e.g. LAION~\cite{Schuhmann2022LAION5B}, WebLI~\cite{chen2022pali}, LVD-142M~\cite{oquab2023dinov2}, etc., \textbf{2)} training objective, e.g. image-text contrastive learning, masked image modelling, classification, etc., and \textbf{3)} model size, e.g. 300M to 4B parameters. Due to this diversity, they incorporate different biases and thus potentially capture different aspects of the underlying depicted scene. 

\begin{table}[tb]
  \caption{\textbf{Overview of vision encoders we benchmarked.} All encoders use a ViT~\cite{Dosovitskiy2020vit} backbone, yet they differ in terms of objective, training data, and model size. \textbf{ITC:} Image-text contrastive learning~\cite{radford2021clip}, \textbf{MIM:} Masked image modelling~\cite{He2021MaskedAA}, \textbf{LGC:} Local-to-global correspondence learning~\cite{caron2021emerging}. See Sec.~\ref{sec:benchmark} for details.
  }
  \centering
\resizebox{0.8\linewidth}{!}{%
\begin{tabular}{cccc}
\toprule
Vision encoder & Parameters & Training data & Objective \\ \midrule
CLIP-L/14~\cite{radford2021clip} & 0.3B & OpenAI WIT & ITC \\ 
OpenCLIP-G/14~\cite{cherti2023reproducible} & 1.8B  & LAION-2B & ITC \\ 
EVA-CLIP-g~\cite{fang2023eva} & 1B  & LAION-400M & MIM~(CLIP features) \\ 
SIGLIP-G/14~\cite{zhai2023sigmoid} & 1.8B & WebLI  & ITC~(Sigmoid loss) \\ 
SILC-G/16~\cite{naeem2023silc} & 1.8B & WebLI & ITC + LGC \\ 
ViT-e~\cite{chen2022pali} & 3.8B & JFT-3B  & Classification  \\ 
ViT-G~\cite{zhai2022scaling} & 1.8B & JFT-3B & Classification \\ 
DINOv2-L/14~\cite{oquab2023dinov2} & 0.3B & LVD-142M & LGC + iBOT~\cite{zhou2021ibot} \\ 
\bottomrule
\end{tabular}}
\label{tab:encoder_stats}

\end{table}

\subsubsection{Pre-training data and objectives.} We pre-train the Q-Former using WebLI~\cite{chen2022pali} dataset at $224 \times 224$ resolution. We use the filtered and de-duplicated English-only subset with 100 million image-text pairs~\cite{chen2022pali,naeem2023silc}. 
We keep both the vision encoder and the LM frozen and only train the parameters of the Q-Former. The model is trained with captioning objective using the alt-text in WebLI as target, as it has been shown to be effective~\cite{xu2023pixel}. Similar to~\cite{li2023blip2}, we remove the last Transformer layer of the encoder. We refer the reader to the supplementary material for additional training details and hyperparameters.

\subsubsection{Evaluation tasks.} We evaluate the obtained VLMs on standard captioning and VQA tasks~(See Fig.~\ref{fig:task_overview} for an overview). 
For captioning, we use the COCO captioning benchmark~\cite{chen2015microsoft} and fine-tune the pre-trained VLM on the commonly employed Karpathy training split~\cite{karpathy2015deep}. Similar to the pre-training, only the Q-Former parameters are updated. For VQA, we again follow the standard practices~\cite{li2023blip2,dai2023instructblip,liu2023improved,chen2022pali} and fine-tune on a mixture of VQA data that enforces the Q-Former to extract more task-aligned visual features. Our mixture includes the training sets of VQAv2~\cite{goyal2017making} and OKVQA~\cite{marino2019ok} as well as the synthetically generated VQA data from VQ$^2$A~\cite{changpinyo2022all}, resulting in total 17M training examples.
We then evaluate the fine-tuned models on VQAv2 and OKVQA validation sets. Similar to~\cite{li2023blip2,dai2023instructblip}, we also evaluate the VLMs' zero-shot VQA capabilities on GQA~\cite{hudson2019gqa}. Finally, we evaluate the zero-shot performance on MMVP~\cite{tong2024eyes} that includes images with semantic differences that are challenging for the current VLMs to handle. 
Unlike captioning, we fine-tune both Q-Former and the LM parameters while keeping the vision encoder frozen, as we find this to improve the performance, similar to~\cite{liu2023improved}. 
For both captioning and VQA tasks, we use $224\times224$ image resolution. See the supplementary for more details and results.

\begin{table}[tb]
  \caption{\textbf{Benchmarking vision encoders on different vision-language tasks}. For COCO captioning, we report CIDEr~\cite{vedantam2015cider} score. For VQA tasks, we report top-1 accuracy. For MMVP, we report average pair accuracy~\cite{tong2024eyes}.  Top-3 best results are shown with \colorbox{blue!20}{Blue},\colorbox{green!20}{Green}, and \colorbox{orange!20}{Orange}, respectively. All VLMs use FlanT5-XL~\cite{chung2022scaling} as the language model and a Q-Former~\cite{li2023blip2} to bridge vision and language modalities. See Sec.~\ref{sec:benchmark} for details.
  }
  \centering
  \resizebox{0.8\linewidth}{!}{%
  \begin{tabular}{cccccc}
\toprule
Vision encoder & \begin{tabular}[c]{@{}c@{}}COCO cap.~$\uparrow$\\ Karpathy val\end{tabular} & \begin{tabular}[c]{@{}c@{}}VQAv2~$\uparrow$\\ Karpathy val\end{tabular} & \begin{tabular}[c]{@{}c@{}}OKVQA~$\uparrow$\\ val\end{tabular} & \begin{tabular}[c]{@{}c@{}}GQA~$\uparrow$\\ test-dev\end{tabular} & \begin{tabular}[c]{@{}c@{}}MMVP~$\uparrow$\\ test\end{tabular}  \\

\midrule

CLIP-L/14   & 133.0               &   74.4     & 61.0  & 48.7  & 15.3              \\ 
OpenCLIP-G  & 128.3                &   73.3      & 60.6  & 48.0  & 22.0             \\ 
EVA-CLIP-g  & \colorbox{green!20}{140.9}               &   \colorbox{blue!20}{77.0}      & \colorbox{green!20}{63.0}  & \colorbox{blue!20}{50.1} & \colorbox{blue!20}{27.3}               \\

SIGLIP-G/14  & 133.0               &   \colorbox{orange!20}{74.7}      & \colorbox{orange!20}{62.5}  & {48.6}  & \colorbox{orange!20}{24.0}               \\ 
SILC-G/16  & \colorbox{blue!20}{141.1}               &   \colorbox{blue!20}{77.0}      & \colorbox{blue!20}{63.4}  & \colorbox{green!20}{49.7} & \colorbox{orange!20}{24.0}                \\ 
ViT-e   & \colorbox{orange!20}{137.8}                  &   \colorbox{green!20}{75.6}      & 61.9  & \colorbox{orange!20}{49.1}  & \colorbox{green!20}{25.3}                \\ 
ViT-G   & 133.8                  &   74.2      & 61.2  & 48.3 & 20.7                 \\ 
DINOv2-L/14    & 127.6            &   71.3      & 59.0  & 48.0 & 22.0                \\ 
\bottomrule
\end{tabular}}\label{tab:benchmark}
\end{table}

\subsubsection{Results.} We report the findings in Table~\ref{tab:benchmark} and make the following observations:
\begin{itemize}
\item \textbf{Encoders with different biases can obtain similar performance.} While we observe some variation in VLMs' performance across different encoders, the gap remains small, especially among the top-performing ones. For example, EVA-CLIP-g, SILC-G/16, and ViT-e achieve the best results for VQAv2, GQA and COCO captioning despite their differences in training objective and data, as shown in Table~\ref{tab:encoder_stats}. Similar observations can be made for OpenCLIP and DINOv2 encoders that have a small performance gap.

\item \textbf{MMVP stays challenging for all encoders.} While it is originally curated with ``CLIP-blind'' pairs in~\cite{tong2024eyes}, i.e. images that CLIP perceives as similar despite their visual differences~(See Fig.~\ref{fig:pull} for examples), most of the encoders stay below the random guess accuracy of $25\%$. Exceptions are EVA-CLIP and ViT-e which slightly surpass the random guess level~($27.3\%$ and $25.3\%$). 

\item \textbf{For tasks requiring compositional reasoning and open-world knowledge, the performance decreases and the gap among VLMs diminishes,} as shown by the results on GQA and OKVQA. More precisely, the standard deviation across all the VLMs' performance on GQA and OKVQA is $0.8$ and $1.36$, respectively, but $1.74$ on VQAv2 and $4.91$ on COCO.
\item \textbf{Scaling the size of vision encoder can improve the performance} as seen by the consistent improvements obtained by ViT-e over ViT-G when scaling from 1.8B to 3.8B parameters.
\item \textbf{Pre-training data distribution can play a critical role.} OpenCLIP-G/14 model is larger than CLIP-L/14 and uses more training data, yet it noticeably underperforms on most of the evaluated VQA and captioning tasks except the MMVP benchmark, where OpenCLIP significantly outperforms the CLIP model. Since both models are trained using the same training objective, this indicates that the impact of the training objective and dataset to the performance of the VLM can be disentangled, and both are important. 
\item \textbf{Performance correlates between VQA and captioning tasks.} Encoders with better performance on VQA tasks generally also perform better on captioning, and vice versa. Spearman's rank correlation between VQA tasks and COCO captioning is $0.90$, $0.80$, $0.79$, $0.56$ for VQAv2, OK-VQA, GQA, and MMVP, respectively. Thus, improving the visual component in VLMs is expected to improve performance for a broad range of tasks.
\end{itemize}

The above results highlight the importance of using the right vision encoder for VLMs, motivating us to focus on the following question: 
\textbf{\textit{Can we broaden the vision capabilities of VLMs through combining vision encoders with different visual biases?} }
We answer positively in the next section.

\section{Combining vision encoders with \method}\label{sec:method}

Motivated by the findings in Sec.~\ref{sec:benchmark}, we propose using multiple vision encoders with different, and potentially complementary, visual biases to create more capable VLMs. To do this, we introduce \method, a method that combines the strengths of different vision encoders while staying efficient in terms of number of trainable parameters. Below, we provide details of \method.

\begin{figure}[t!]
  \centering
  \includegraphics[width=\textwidth]{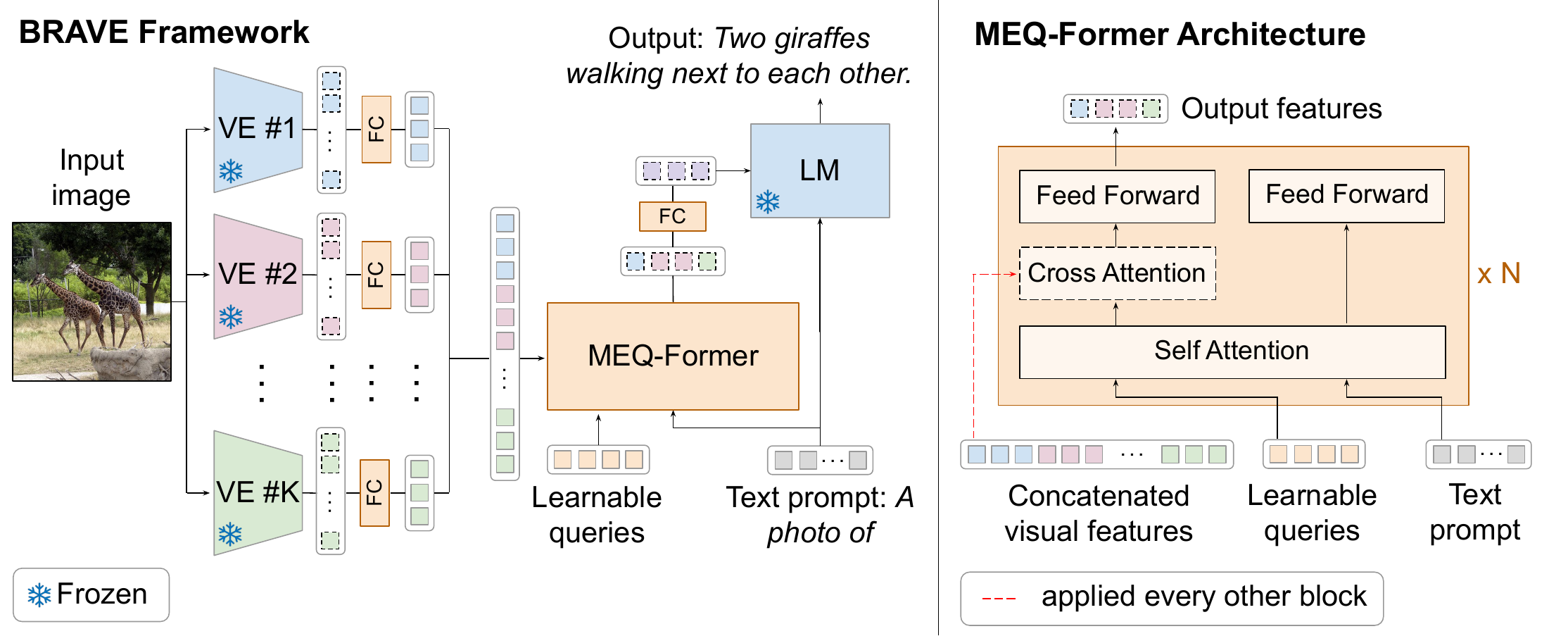}
  \caption{\textbf{Overview of \method.} \textbf{Left:} We keep all the vision encoders~(VEs) and the language model~(LM) frozen.
  The linear projection layers are used to concatenate features from $K$ different VEs, e.g. $K=5$, sequence-wise. These are then resampled by the \bridge which accepts a set of learnable queries and a text prompt describing the task as inputs. The output of \bridge is projected to the input space of the LM using fully-connected~(FC) layers. The total number of trainable parameters is 116M~($\approx1\%$ of the total parameters). \textbf{Right:} Architecture of the \bridge with $N=12$ transformer layers. It interacts with the concatenated visual features through cross-attention layers and produces a fixed-length output to be fed as soft visual prompt to the frozen LM.}
  \label{fig:method}
\end{figure}

\subsubsection{Overview.} As depicted in Fig.~\ref{fig:method}, we introduce the \textit{\textbf{m}ulti-\textbf{e}ncoder \textbf{q}uerying trans\textbf{former}}, or \bridge, that combines visual features from an \textit{arbitrary} set of encoders and outputs a \textit{fixed-length} visual representation that is given as a soft prompt to a frozen LM. It takes as an input a sequence comprised of a fixed number of learnable queries and embedded text tokens which describe the task -- for captioning, they are the textual prompt \textit{``A photo of''}, while for VQA, they are the question prompt for a given image~(See supplementary for examples). 
\bridge interacts with visual features through cross-attention layers. The visual features from different encoders are first linearly projected to the same dimension and then concatenated sequence-wise. The obtained concatenated feature is given as a key and value pair to the cross-attention layers in \bridge, and is cross-attended to by the \bridge's query sequence~(Fig.~\ref{fig:method}). This resampling enables to efficiently processing a large number of visual features since it bypasses the quadratic complexity of self-attention. It also acts as a ``bottleneck'' that keeps the total number of VLM's parameters low compared to a naive ensembling of VLMs that is prohibitively expensive~(See a comparison in Sec.~\ref{sec:ablations}). 
Moreover, the visual features are not ``marked'' with any encoder-specific embedding, thus \bridge is not biased to differentiate between encoders, allowing for a simpler design. 

\subsubsection{Differences with previous work.} Our work differs from other popular approaches like LLaVa~\cite{liu2023improved} and PaLI~\cite{chen2022pali} that require significantly more trainable parameters~($\approx$10B) and do not have a resampling mechanism. Resampling visual features from a single encoder has been successfully demonstrated in previous work for single-\cite{li2023blip2,dai2023instructblip} and multiple-frames~\cite{korbar2023text,yu2023self, Alayrac2022Flamingo}. To the best of our knowledge, our work is the first one demonstrating an effective resampling mechanism to consolidate visual features from \textit{several arbitrarily different} encoders while keeping the number of trainable parameters small~(our \bridge having 116M parameters vs BLIP-2's Q-Former is 188M). See Sec.~\ref{sec:related_work} for more comparisons.

\subsubsection{Pre-training details.} We use the same data and objective as in Sec.~\ref{sec:benchmark} and only train the \bridge while keeping the vision encoders and the LM frozen. During pre-training, we randomly mask features from each encoder with $20\%$ probability as we found that this can act as a regularizer and enforces \bridge to avoid local minima by attending only to the features of a single encoder~(Ablated in Sec.~\ref{sec:ablations}). After pre-training, we fine-tune the \bridge and (optionally) the LM on the downstream tasks, as explained in Sec.~\ref{sec:results}.

\subsubsection{Architecture.} For the main results, we combine five vision encoders, namely \textit{EVA-CLIP-g}~\cite{fang2023eva}, \textit{CLIP-L/14}~\cite{radford2021clip}, \textit{SILC-G/16}~\cite{naeem2023silc}, \textit{ViT-e}~\cite{chen2022pali}, and \textit{DINOv2-L/14}~\cite{oquab2023dinov2} to cover all training datasets and objectives from Table~\ref{tab:encoder_stats}. Please see Sec.~\ref{sec:ablations} and the supplementary for analysis on the contributions of encoders.
For the LM, we employ FlanT5-XL~\cite{chung2022scaling} as before. We set the visual feature dimension after linear projection as $1408$ for concatenation. For \bridge, we use $32\times5=160$ learnable queries to proportionally scale the model capacity to the number of vision encoders and set the hidden dimension as $768$. The \bridge resampling results in $14\times$ reduction in total feature size (from $1223\times 1408$ to $160\times 768$). The total number of trainable parameters during pre-training is 116M which is about $1\%$ of the total number of parameters in the VLM~(10.3B).

\subsubsection{Implementation.} We use FLAX~\cite{heek1flax} with the Scenic~\cite{dehghani2022scenic} library to implement our models and training pipeline. For evaluations, we use both Scenic and EvalAI API~\cite{EvalAI}. We pre-train with batch size of 1024 on 64 TPUv5 chips. Please see the supplementary for more details. 

\section{Experiments}\label{sec:results}

\begin{figure}[t!]
  \centering
  \includegraphics[width=\textwidth]{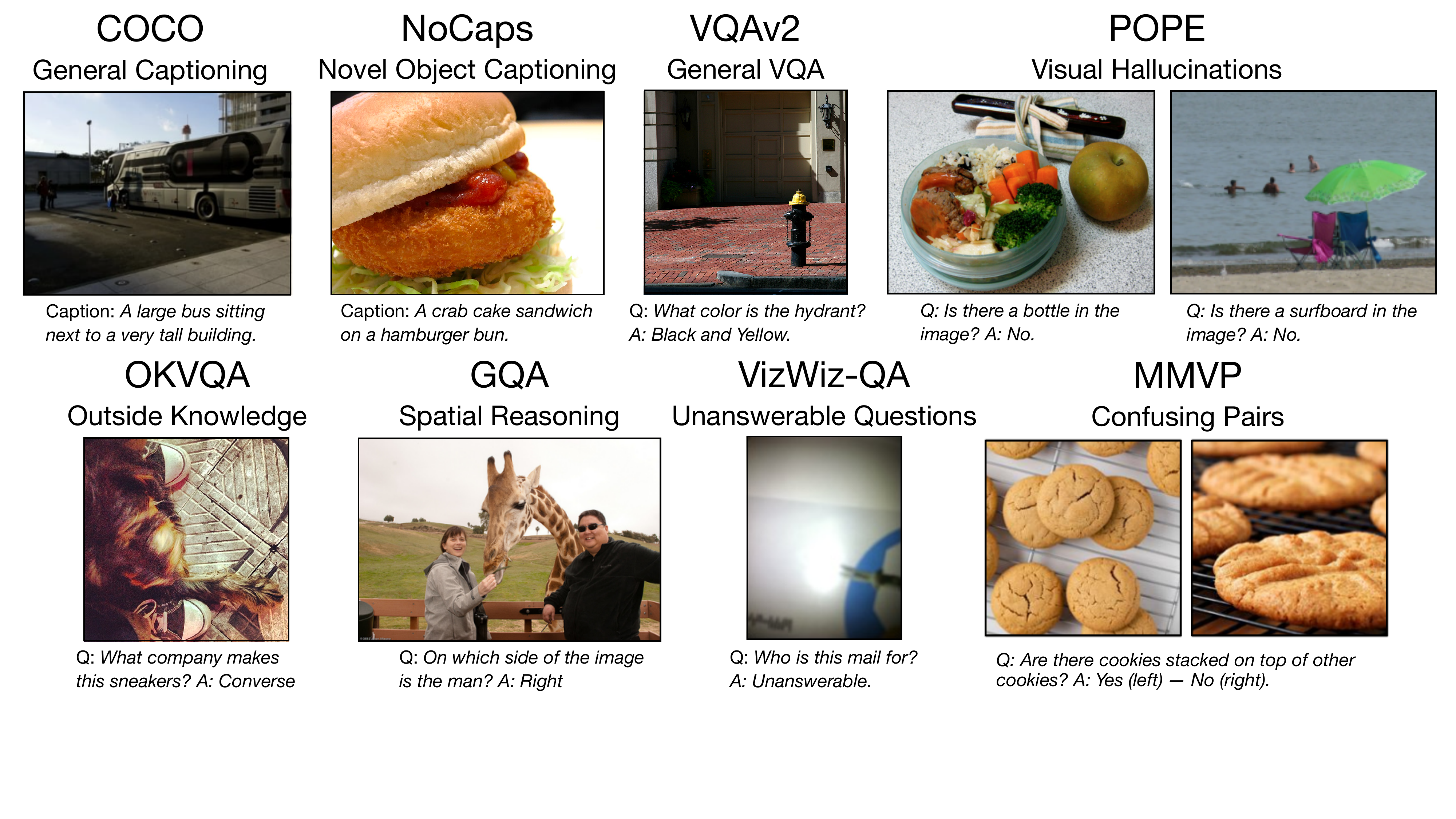}
  \vspace{-0.6in}
  \caption{Overview of the evaluation tasks. They evaluate different capabilities of VLMs, which is important to understand their strengths and weaknesses. The visualizations are obtained from the corresponding publications~\cite{chen2015microsoft,agrawal2019nocaps,goyal2017making,li2023evaluating,marino2019ok,hudson2019gqa,gurari2018vizwiz,tong2024eyes}, respectively.
  }
  \label{fig:task_overview}
\end{figure}

\subsection{Overview}
We perform evaluations on a broad range of captioning~(Sec.~\ref{sec:captioning}) and VQA tasks~(Sec.~\ref{sec:vqa}) to show the effectiveness of \method. Please see Fig.~\ref{fig:pull} right for a summary of the results and Fig.~\ref{fig:task_overview} for an overview of the tasks. We then provide a comprehensive analysis of \method in Sec.~\ref{sec:ablations}. 

For captioning, we evaluate on the popular COCO~\cite{chen2015microsoft} and NoCaps~\cite{agrawal2019nocaps} benchmarks. The latter includes out-of-distribution samples with novel classes that do not exist in the COCO benchmark, hence it serves as a challenging testbed to assess robustness. The results~(Table~\ref{tab:captioning}) show that \method pushes the state-of-the-art for both benchmarks while staying efficient in terms of trainable parameters.  

{For VQA, we evaluate on the popular VQAv2~\cite{goyal2017making}, OKVQA~\cite{marino2019ok}, GQA~\cite{hudson2019gqa}, and VizWiz-QA~\cite{gurari2018vizwiz} benchmarks. To assess the robustness of VLMs trained with \method, we additionally include 1) the POPE~\cite{li2023evaluating} benchmark that measures object hallucinations in VLMs, and 2) the MMVP~\cite{tong2024eyes} benchmark which exposes the shortcomings of CLIP-based vision encoders that are widely used in VLMs. Our results~(Table~\ref{tab:vqa}) show that \method achieves consistent improvements over recent methods for all the tasks by leveraging diverse visual features.}

\subsection{Image captioning}\label{sec:captioning} 

\subsubsection{Fine-tuning details.} Similar to Sec.~\ref{sec:benchmark}, we only fine-tune the \bridge on COCO captioning and keep the vision encoders and the LM frozen. During fine-tuning, we set a higher input resolution as $336 \times 336$ to further boost the performance, similar to previous works~\cite{chen2022pali,li2023blip2,liu2023improved}~(See Sec.~\ref{sec:ablations} for an ablation). Following~\cite{li2023blip2,dai2023instructblip, Yu2022CoCa,wang2022git,chen2022pali}, we also zero-shot evaluated the COCO fine-tuned model on NoCaps~\cite{agrawal2019nocaps}. Please see the supplementary for more training details.

\subsubsection{Results.}

We summarize the captioning evaluations in Table~\ref{tab:captioning}. \method uses the least number of trainable parameters~(116M) yet achieves strong results for both COCO and NoCaps benchmarks. For NoCaps, \method is the best performing method with significant gains over recent methods. This is especially the case for out-domain samples with novel classes, demonstrating the usefulness of diversity in visual features for robustness. For COCO, \method stays competitive with the best performing model, PaLI-17B~\cite{chen2022pali}. Notably, \method achieves this while using $150\times$ fewer trainable parameters~(116M vs 16.9B), $16\times$ less pre-training data~(100M vs 1.6B) and $3\times$ fewer image pixels~($336\times 336$ vs $588\times 588$) than PaLI-17B, suggesting that having different visual biases is effective for generalization, while keeping the sample complexity low.

\begin{table}[tb]
  \caption{\textbf{Captioning.} We compare \method with state-of-the-art captioning methods on COCO and NoCaps~(zero-shot). CIDEr~\cite{vedantam2015cider} score is reported~(Best in \textbf{bold}, second best is \underline{underlined}). Numbers of other methods are taken from the corresponding publications. Dashed line means no result was reported. \method achieves the best results in NoCaps evaluation sets~(both out-domain and overall in validation \& test) and second best for COCO, while having much fewer trainable parameters than other methods. For example, compared to PaLI-17B~\cite{chen2022pali}, \method uses significantly fewer trainable parameters, training data, and image pixels. 
  Yet, it outperforms the baselines on NoCaps and stays competitive on COCO, demonstrating the effectiveness of having different visual biases for generalization. See Sec.~\ref{sec:captioning} for more details.
  }
  \centering
  \resizebox{\linewidth}{!}{
\begin{tabular}{cccccccc}
\toprule
                              & \multicolumn{2}{c}{\# params} & COCO (fine-tuned) & \multicolumn{2}{c}{NoCaps (zero-shot, val)} & \multicolumn{2}{c}{NoCaps (zero-shot, test)} \\                    
                              \cline{2-8}
\multicolumn{1}{c}{Method}   & Trainable          & Total         & Karpathy test       & out-domain   & overall    & out-domain  & overall \\ \midrule
\multicolumn{1}{c}{Flamingo~\cite{Alayrac2022Flamingo}} & 10.6B          & 80B           & 138.1              & -            & -         & -                    & -       \\
\multicolumn{1}{c}{SimVLM~\cite{wang2021simvlm}} & 632M          & 632M           & 143.3             & 113.7                  & 112.2                 & -           & 110.3       \\ 
\multicolumn{1}{c}{Qwen-VL~\cite{bai2023qwen}}   & 9.6B           & 9.6B          & -                & -        & 121.4               & -           & -       \\ 
\multicolumn{1}{c}{BLIP-2~\cite{li2023blip2}}   & 1.1B           & 4.1B          & 144.5                    & 124.8        & 121.6              & -           & -       \\ 

\multicolumn{1}{c}{InstructBLIP~\cite{dai2023instructblip}}   & {188M}           & 14.2B          & -                 & -        & 121.9            & -           & -       \\ 
\multicolumn{1}{c}{CoCa~\cite{Yu2022CoCa}}     & 2.1B           & {2.1B}          & 143.6                     & -            & 122.4             & -           & 120.6   \\ 

\multicolumn{1}{c}{GiT2~\cite{wang2022git}}     & 5.1B           & 5.1B          & 145.0                   & \underline{130.6}        & {126.9}         & 122.3       & \underline{124.8}   \\ 
\multicolumn{1}{c}{PaLI-17B~\cite{chen2022pali}} & 16.9B            & 16.9B         & \textbf{149.1}                  & -            & \underline{127.0}          & \underline{126.7}       & {124.4}   \\ 
\multicolumn{1}{c}{\method}     & \textbf{116M}             & 10.3B         & \underline{148.0}                 &  \textbf{133.3}            &   \textbf{127.6}                  &    \textbf{127.1}         &  \textbf{125.6}       \\ \bottomrule
\end{tabular}
}\label{tab:captioning}
\end{table}

\subsection{Visual question answering}\label{sec:vqa}

\subsubsection{Fine-tuning details.} We use the same VQA mixture as in Sec.~\ref{sec:benchmark} and fine-tune both the \bridge and the LM at $224\times 224$ resolution. Vision encoders are kept frozen. This is followed by a high-resolution fine-tuning stage on VQAv2 and OKVQA training sets at $336\times336$ resolution, similar to previous work~\cite{chen2022pali,liu2023improved,li2023blip2}. The resulting model is evaluated on VQAv2 and OKVQA benchmarks as well as zero-shot evaluated on GQA~\cite{hudson2019gqa}, VizWiz-QA~\cite{gurari2018vizwiz}, MMVP~\cite{tong2024eyes} and POPE~\cite{li2023evaluating}, following~\cite{li2023blip2,dai2023instructblip,liu2023improved,tong2024eyes}. To be able to compare to the methods that fine-tune on GQA training set, e.g.~\cite{liu2023improved,bai2023qwen, lin2023sphinx}, we also perform an additional training stage on top of the mixture-finetuned model, and evaluate the model on the GQA evaluation set~(test-dev). Please see the supplementary for the details.

\subsubsection{Results.}

\begin{figure}[t!]
  \centering
  \includegraphics[width=\textwidth]{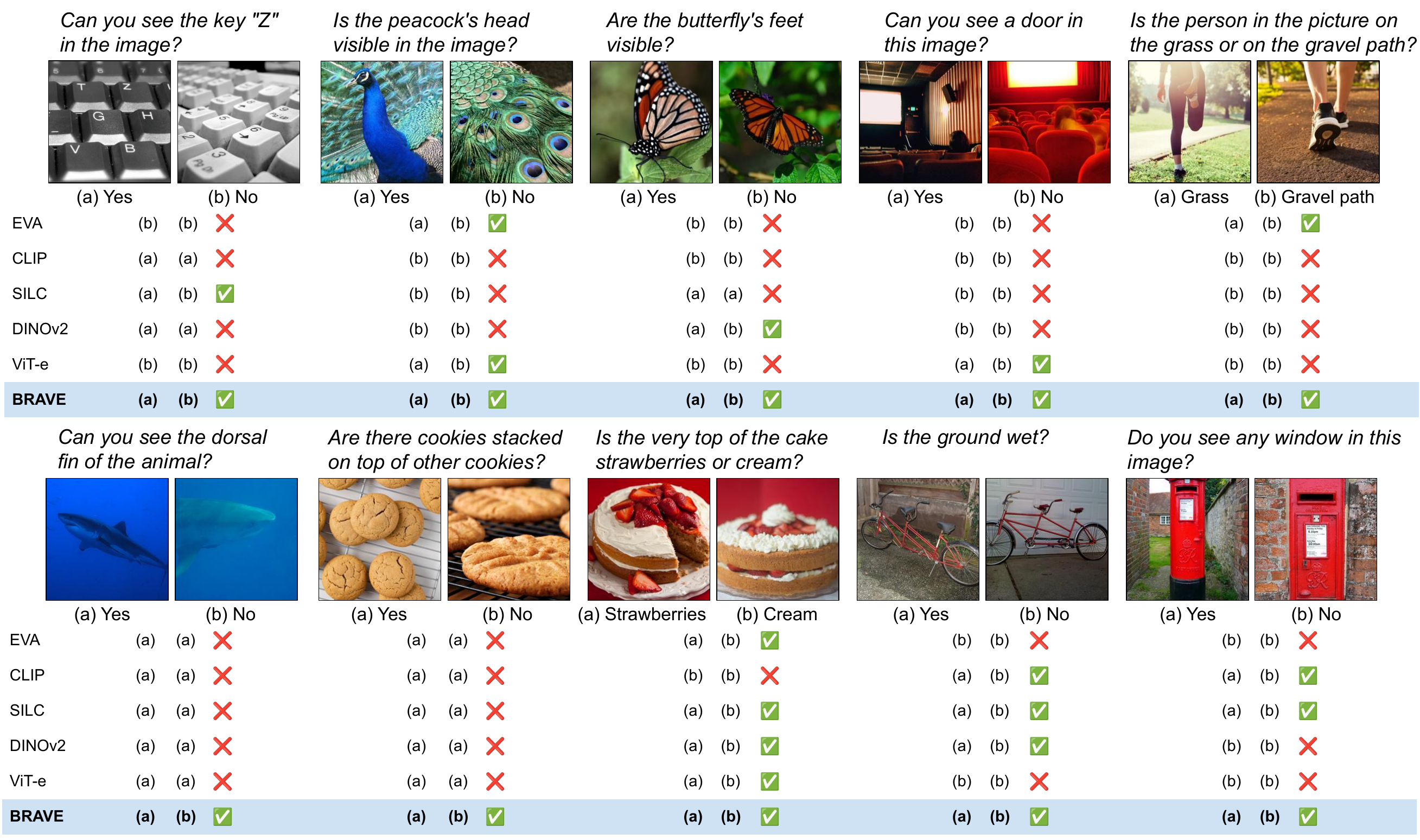}
  \caption{
  {\textbf{Qualitative results.} 
  We compare predictions of \method and the VLMs with different vision encoders, e.g. CLIP, on samples from the MMVP benchmark. Following~\cite{tong2024eyes}, a model is considered correct only if it answers to both images in a pair correctly, i.e. if it can successfully differentiate between images with semantic differences. Note that the images in a pair are seen independently, i.e. neither of the images is provided as context for the other one. All encoders output some correct predictions, yet none of them performs consistently well on a broad range of inputs. \method alleviates this by combining diverse visual features, leading to a more consistent performance. The quantitative difference is indeed stark: $42\%$ for \method vs $27.3\%$ for best single encoder~(Tables~\ref{tab:benchmark} and \ref{tab:vqa}). See supplementary for more qualitative results.}
  }
  \label{fig:qualitative}
\end{figure}

We summarize the VQA evaluations in Table~\ref{tab:vqa}. \method achieves strong results for all the benchmarks in both fine-tuned and zero-shot evaluation scenarios. For OKVQA, GQA, VizWiz-QA, MMVP, and POPE benchmarks, \method achieves the best results, and for VQAv2, it stays competitive with PaLI-17B:

\begin{itemize}
    \item For MMVP, \method significantly improves the performance over previous approaches, making it less susceptible to failure modes of CLIP-based encoders. This can be further seen from the qualitatives in Figures~\ref{fig:pull} and \ref{fig:qualitative}.
    \item For VizWiz-QA, \method achieves the best results without using any scene text understanding or OCR data during pre-training, as opposed to~\cite{liu2023improved, chen2023minigpt, bai2023qwen}.
    \item For GQA and OKVQA that requires different capabilities, i.e. spatial reasoning and open-world knowledge, \method noticeably improves the performance.
    \item For POPE, \method further reduces the visual hallucinations over dual encoder methods~\cite{lin2023sphinx, tong2024eyes}, confirming that using multiple encoders can be effective for visual grounding~(Also see Sec.~\ref{sec:related_work} for a discussion).
    \item For VQAv2, \method is the second best after PaLI-17B, while staying more efficient in terms of training data, model size, and input resolution, similar to the COCO captioning result in Sec.~\ref{sec:captioning}. We expect \method to benefit from such scaling efforts as well, which is a future direction to investigate.
\end{itemize}

\begin{table}[tb]
  \caption{\textbf{Visual question answering.} We compare \method with state-of-the-art VQA methods on VQAv2, OKVQA, GQA, VizWiz-QA, MMVP, and POPE. Top-1 accuracies are reported~(Best in \textbf{bold}, second best is \underline{underlined}). For MMVP, we report average pair accuracy~\cite{tong2024eyes}. Numbers of other methods are taken from the corresponding publications. Dashed line means no result was reported. \method achieves the best results for six out of seven benchmarks, while staying efficient in terms of trainable parameters. Similar to the COCO captioning results in Table~\ref{tab:captioning}, \method stays competitive with PaLI-17B~\cite{chen2022pali} for VQAv2, while using significantly less pre-training data, image resolution, and model parameters~(both trainable and total). See Sec.~\ref{sec:vqa} for details. 
  }
  \centering
  \resizebox{\linewidth}{!}{
\begin{tabular}{cccccc|cccc}
\toprule
\multicolumn{1}{l}{}                 & \multicolumn{2}{c}{\# params} &  \multicolumn{3}{c}{Fine-tuned}                    & \multicolumn{4}{c}{Zero-shot} \\ 
\cline{2-10}
Method          & Trainable & Total  &  \begin{tabular}[c]{@{}c@{}}VQAv2\\ test-dev\end{tabular} & \begin{tabular}[c]{@{}c@{}}OKVQA\\ val\end{tabular}  & \begin{tabular}[c]{@{}c@{}}GQA\\ test-dev\end{tabular} & \begin{tabular}[c]{@{}c@{}}VizWiz-QA\\ test-dev\end{tabular}     & \begin{tabular}[c]{@{}c@{}}GQA\\ test-dev\end{tabular}   & \begin{tabular}[c]{@{}c@{}}MMVP\\ test\end{tabular}    & \begin{tabular}[c]{@{}c@{}}POPE\\ test\end{tabular}    \\ \midrule
SimVLM~\cite{wang2021simvlm}        & 632M    & 632M            & 80.0             & -  & -  &-           & -     & -     &  -  \\ 
Flamingo~\cite{Alayrac2022Flamingo}        & 10.2B    & 80B            & 82.0             & 57.8  & -      &31.6      & -     & -     &  -  \\
MiniGPT-v2~\cite{chen2023minigpt}        & 7B    & 8B            & -             & 57.8  & 60.1    &\underline{{\color{black}53.6}}        & -     & -     & -   \\ 

GiT2 ~\cite{wang2022git}           & 5.1B     & {5.1B}         & 81.7             & -  & -          &-      & -     & -    & -    \\ 
Qwen-VL~\cite{bai2023qwen}            & 9.6B     & 9.6B        & 79.5             & 58.6  & 59.3       &{\color{black}35.2}         & -     & -    &  -   \\ 
SPHINX-2k~\cite{lin2023sphinx}            & 13B     & 16.5B        & 80.7             & 62.6  & 63.1      &{\color{black}44.9}          & -     & -   & \underline{87.2}      \\ 
PaLI-17B~\cite{chen2022pali}        & 16.9B      & 16.9B          & \textbf{84.3}             & \underline{64.5}   & -  &-         & -     & -     & -    \\
BLIP-2~\cite{li2023blip2}          & {1.2B}     & 12.1B        & 81.6    & 54.7  &- &29.4   & 44.7  & -    & 85.3     \\ 
InstructBLIP~\cite{dai2023instructblip}    & {188M}     & 14.2B        & -                & 55.5  & -  & {\color{black}33.4} & \underline{49.5}  & 16.7  & 78.9    \\ 
LLaVa$^{1.5}$~\cite{liu2023improved}       & 13B      & 13.4B          & 80.0             & -        & \underline{63.3}     &\underline{{\color{black}53.6}}    & -     & 24.7  & 85.9    \\ 
LLaVA$^{1.5}$~(I-MoF)~\cite{tong2024eyes} & 13B      & 13.6B          & 79.3        & -            & -   &-  & -     & \underline{31.3} & 86.7 \\
\method         & {3B}       & {10.3B}    & \underline{82.5}                & \textbf{66.0}         & \textbf{66.3}     &\textbf{54.2} & \textbf{52.7}  & \textbf{42.0}   & \textbf{87.6}   \\ \bottomrule
\end{tabular}
}\label{tab:vqa}
\end{table}

\subsection{Analysis of \method}~\label{sec:ablations}

We study different aspects of \method including an ablation study, contribution of vision encoders to the final performance, comparing \bridge to an ensembling strategy, and the role of pre-training data. Please see the supplementary for more qualitative and quantitative results.

\subsubsection{Ablation study.} We perform a comprehensive study in Table~\ref{tab:ablation} where we ablate different design choices, e.g. training data and choice of language model. The ablations are labeled from A1 to A8, each showing a particular deviation from the original pre-training setup A0 explained in Sec.~\ref{sec:method}. We summarize the key findings:

\begin{itemize}
    \item \textbf{(A1):} LM fine-tuning for VQA significantly boosts the performance compared to frozen LM, similar to the observation in~\cite{liu2023improved}. To minimize the performance loss, we perform a LoRA~\cite{Hu2021LoRA} fine-tuning by inserting LoRA layers to the LM. With a rank of $128$, we compensated for $70\%$ of the performance gap while using $10\times$ fewer trainable parameters than full fine-tuning~(324M vs 3B). This underscores the development of parameter-efficient fine-tuning methods as an important direction for more efficient VLMs.
    \item \textbf{(A2):} Using synthetic VQA data~\cite{changpinyo2022all} noticeably enhances our performance, while being much cheaper to collect than human annotated data. This can potentially be further improved by using more diverse templates~\cite{guo2023images,tiong2022plug}.
    \item \textbf{(A3,A4,A5):} Encoder dropout during training improves the performance for COCO captioning and VQAv2 but slightly degrades for OKVQA. Using text prompt as an additional input to \bridge helps with extracting better task-aligned features. Similarly, high-resolution fine-tuning stage gives a boost to the performance at the expense of processing more visual tokens.
    \item \textbf{(A6,A7):} These ablations combine the impact of previously ablated design choices, further demonstrating their contribution to the final performance. 
    \item \textbf{(A8):} A stronger LM significantly helps for all the tasks, underscoring that scaling the VLM along the vision and language axes simultaneously have complementary benefits, and both are important.
\end{itemize}

\begin{table}[t]
  \caption{ \textbf{Ablation study of \method.} We start from the base pre-training setup explained in Sec.~\ref{sec:method}, denoted as A0. Each row corresponds to a particular deviation from A0, while keeping the rest fixed. For example, A5 evaluates the performance when high-resolution fine-tuning is disabled. N/A means the ablation is not applicable for that evaluation. See Sec.~\ref{sec:ablations} for the discussions and supplementary for more results.
  }
  \label{tab:ablation}
  \centering
  \resizebox{0.95\linewidth}{!}{%
\begin{tabular}{ccccccc}
\toprule
Ablation ID & Study Subject                  & Original                         & Change                                                                                           & \begin{tabular}[c]{@{}c@{}}COCO Captioning\\ (Karpathy val)\end{tabular} & \begin{tabular}[c]{@{}c@{}}VQAv2 \\ (Karpathy val)\end{tabular} & \begin{tabular}[c]{@{}c@{}}OKVQA \\ (val)\end{tabular} \\ \midrule
A0 & Base           & -                              & -                                                                                                  & 147.0                                                                       & 81.8                                                              & 65.7                                                      \\ \midrule
A1 & \makecell{LM fine-tuning \\ for VQA}         & \makecell{Full \\ fine-tuning}   & \begin{tabular}[c]{@{}c@{}}No fine-tuning\\ LoRA (r=64) \\ LoRA (r=128)\end{tabular} & \begin{tabular}[c]{@{}c@{}}N/A\\ N/A\\ N/A\end{tabular}                       & \begin{tabular}[c]{@{}c@{}}78.6\\ 80.7\\ 81.0 \end{tabular}              & \begin{tabular}[c]{@{}c@{}}57.5\\ 62.8\\ 62.9\end{tabular}      \\ \midrule

A2 &Synthetic VQA data~\cite{changpinyo2022all}               & \cmark & \xmark                                                                                              & N/A                                                                       & 81.1                                                              & 64.0                                                     \\ \midrule
A3 & Encoder dropout        & \cmark                      & \xmark                                                                                             & 145.3                                                                       & 81.3                                                              & 66.0                                                      \\ \midrule
A4 & \bridge text input             & \cmark & \xmark                                                                         & 145.9                                                                       & 81.4                                                              & 64.9                                                      \\ \midrule
A5 & High-res fine-tuning & \cmark                      & \xmark                                                                                          & 145.2                                                                       & 79.6                                                              & 65.0                                                      \\ \midrule
A6 & A3 + A5 & \cmark                      & \xmark                                                                                          & 144.0                                                                       & 79.0                                                              & 64.5                                                      \\ \midrule
A7 & A4 + A5 & \cmark                      & \xmark                                                                                          & 145.1                                                                       & 78.3                                                              & 63.4                                                      \\ \midrule
A8 & Language model         & FlanT5-XL                      & FlanT5-L                                                                                           & 142.5                                                                       & 79.9                                                              & 65.5                                                      \\ \bottomrule
\end{tabular}
}
\end{table}

\begin{figure}[t]
  \centering
  \includegraphics[width=\textwidth]{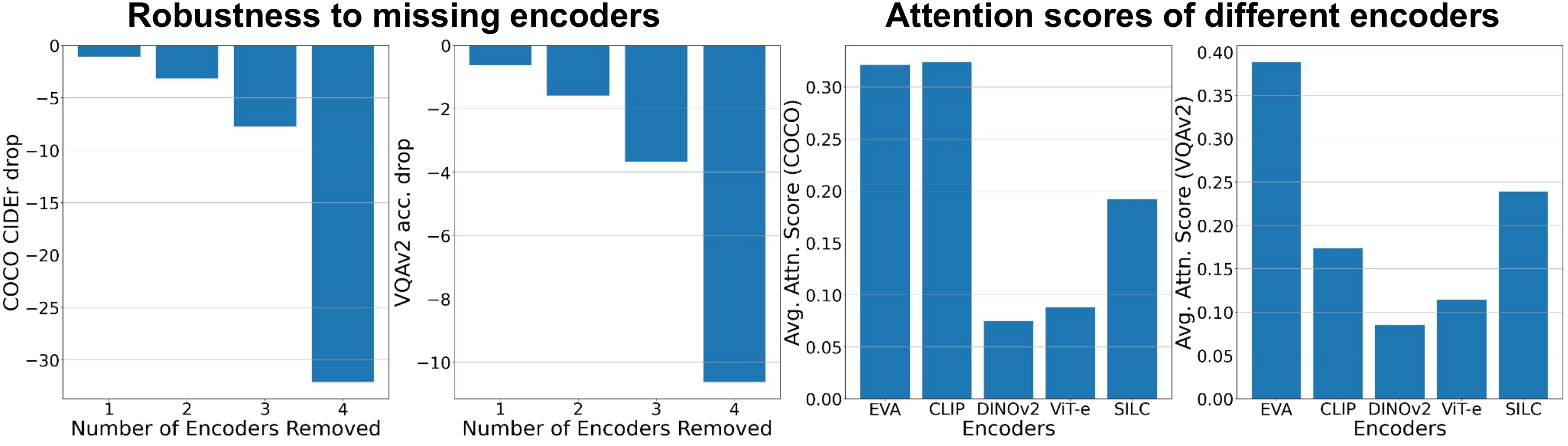}
  \caption{\textbf{Contribution of vision encoders to \method.} \textbf{Left:} We analyze the robustness of \method when a subset of encoders are removed during evaluation. We report the average drop in CIDEr for COCO and accuracy for VQAv2. \textbf{Right:} We compute average attention scores for different vision encoders cross-attended by the \bridge for COCO and VQAv2.
  See Sec.~\ref{sec:ablations} for the discussions. 
  }
  \label{fig:robustness}
\end{figure}

\subsubsection{Contribution of vision encoders.} To understand the impact of vision encoders to the final performance of \method, we perform additional evaluations:
\begin{itemize}
    \item In Fig.~\ref{fig:robustness}-left, we perform a robustness analysis by removing a subset of vision encoders during inference. This is performed for all the combinations and we report the average drop in performance for COCO captioning and VQAv2 tasks~(Full results in supplementary). 
The results show that the performance of \method degrades gracefully up to two encoder removals for both tasks, indicating some redundancy among the encoders. Because of this, even if some encoders are missing, \method can make use of the remaining set of encoders to compensate. 
Beyond the two encoder removal, the drop in the performance is more severe, suggesting that some features are more unique and consequently harder to replace using the remaining set. 

\item In Fig.~\ref{fig:robustness}-right, we compute attention scores of features from different encoders that are cross-attended by the \bridge queries. These are averaged for randomly sampled 1k samples from COCO and VQAv2 validation sets. The results show that the \bridge can cross-attend, or ``pay attention to'', some encoders more than the others depending on the downstream task adaptively, further confirming the usefulness of having different biases. 
\end{itemize}

\begin{table}[ht]
\caption{ \textbf{\bridge vs Ensembling.} We compare resampling vision encoder features by using an ensemble of Q-Formers and the proposed \bridge. The latter uses significantly fewer trainable parameters and better captures the strengths of different vision encoders, leading to consistently better performance. All evaluations are performed at $224\times224$ resolution. See Sec.~\ref{sec:ablations} for details.
  }
  \label{tab:ensemble}
\centering
  \resizebox{0.8\linewidth}{!}{
\begin{tabular}{cccccc}
\toprule
Bridge                       & \# of parameters & COCO Cap. & VQAv2  & OKVQA & GQA \\ \midrule
Q-Former Ensemble  & 605M                  & 140.9                  & 78.5         & 64.3        & 50.6      \\ 
\bridge      & \textbf{116M}                    & \textbf{145.2}                   & \textbf{79.6}         & \textbf{65.0}        & \textbf{51.5}      \\ \bottomrule
\end{tabular}
}
\end{table}

\subsubsection{\bridge vs Ensembling.} We show that resampling the features from all vision encoders with \bridge lead to strong performance while being efficient in terms of the number of trainable parameters. In Table~\ref{tab:ensemble}, we compare \bridge to an ensemble of Q-Formers~\cite{li2023blip2}. Each Q-Former is first fully pre-trained with its corresponding single vision encoder. This is followed by a joint pre-training and a fine-tuning stage by ensembling all vision encoders and their pre-trained Q-Formers. The outputs of all Q-Formers are fed as input to the same LM~(FlanT5-XL~\cite{chung2022scaling}). We use the same five encoders as before, hence the only difference is using an ensemble of Q-Formers instead of \bridge for resampling, resulting in $5\times$ the number of trainable parameters. Our results in Table~\ref{tab:ensemble} show that \bridge resamples visual features more effectively for several tasks while using fewer trainable parameters.

\begin{table}[ht]
\caption{ \textbf{Role of pre-training data.} We compare VLMs pre-trained on WebLI~\cite{chen2022pali} and CC3M~\cite{Changpinyo2021CC12M} datasets. All evaluations are performed at $224 \times 224$ resolution. The latter has significantly less image-text pairs which leads to a degradation in the performance, suggesting more studies are needed to reduce the sample complexity of VLM training. See Sec.~\ref{sec:ablations} for details.
  }
  \label{tab:pretraining}
\centering
  \resizebox{0.6\linewidth}{!}{%
\begin{tabular}{cccccc}
\toprule
Pre-training Dataset & COCO Cap. & VQAv2 & OKVQA & GQA \\ \midrule
CC3M                 &    138.3                     &  76.9            &      63.4       &    50.0       \\ 
WebLI                & \textbf{145.2}                   & \textbf{79.6}         & \textbf{65.0}        & \textbf{51.5}      \\ \bottomrule
\end{tabular}
}
\end{table}

\subsubsection{Role of pre-training data.} In Table~\ref{tab:pretraining} we evaluate the impact of pre-training data by comparing the \method models pre-trained on WebLI~\cite{chen2022pali} and CC3M~\cite{Changpinyo2021CC12M} datasets. The latter has about $30\times$ fewer samples than the former, and pre-training with it leads to a noticeable degradation in performance, suggesting that more work is needed to reduce the sample complexity of VLMs, e.g. as studied in~\cite{jian2024bootstrapping}.

\section{Related work} \label{sec:related_work}

Several works have focused on training VLMs to solve tasks requiring both vision and text understanding capabilities. Most of these works make use of pre-trained vision encoders~\cite{radford2021clip,Fang2022EVA, zhai2022scaling, Dosovitskiy2020vit, brock2021high, ilharco_gabriel_2021_5143773} and LMs~\cite{Raffel2019T5, chung2022scaling, chiang2023vicuna, touvron2023llama, zhang2022opt, hoffmann2022training,bai2023qwen}. Architecture-wise, different approaches were developed to project visual features to the textual embedding space, e.g. Q-Former\cite{li2023blip2,dai2023instructblip,zhu2023minigpt}, P-Former~\cite{jian2024bootstrapping}, cross-attention~\cite{bai2023qwen, chen2022pali, Alayrac2022Flamingo,ye2023mplug}, linear projector~\cite{liu2023visual,chen2023shikra,chen2023minigpt} and MLP~\cite{liu2023improved, wang2023cogvlm}. Different objectives were also considered depending on the downstream tasks, e.g. image-text contrastive learning~\cite{radford2021clip,jia2021scaling,naeem2023silc,zhai2022lit,yao2021filip}, auto-regressive text generation~\cite{cho2021unifying, Chen2021Pix2seqAL, wang2022ofa, chen2022pali, wang2022git, wang2021simvlm} or both~\cite{Yu2022CoCa, li2021align, li2022blip, li2023blip2}. Please also see~\cite{zhang2024mm} for a recent survey. In contrast to ours, these works consider a single vision encoder which has inherent limitations~\cite{li2023evaluating, huang2024visual, thrush2022winoground, tong2024eyes}~(as shown in Sec.~\ref{sec:benchmark}). 

Concurrent to our work, LLaVa-MoF~\cite{tong2024eyes} proposes combining CLIP~\cite{radford2021clip} and DINOv2~\cite{oquab2023dinov2} encoders using separate adapters. SPHINX~\cite{lin2023sphinx} mixes CLIP, DINOv2, and Q-Former outputs with linear projection layers. {Both of the methods input a concatenation of the extracted visual embeddings to the LM, which is not scalable to multiple encoders.} We differ from these works by proposing a flexible and unified resampling mechanism that combines features from several vision encoders while staying efficient and improving the performance~(Tables~\ref{tab:captioning},~\ref{tab:vqa}).

Our work can also be viewed as scaling the VLMs along the vision axis, unlike other works that explored scaling homogenously model size across the visual and/or language component~\cite{team2023gemini, achiam2023gpt, Alayrac2022Flamingo, awadalla2023openflamingo, gong2023multimodal, chen2022pali,chen2023pali, driess2023palm, Wang2022BEiT3 }, training data~\cite{bai2023qwen, wang2022git, jia2021scaling, ilharco_gabriel_2021_5143773, schuhmann2022laion, kakaobrain2022coyo-700m, Changpinyo2021CC12M, lin2023vila, wang2023cogvlm} and number of modalities~\cite{4m, Lu2022UnifiedIOAU, lu2023unified, bai2023sequential,Liu2023Prismer, han2023imagebind,yuan2021florence}. 
{\method can be combined with these efforts to further improve performance. } Moreover, several studies investigated the visual aspect of VLMs by evaluating them against hallucinations and other robustness issues~\cite{huang2024visual, liu2024survey, liu2023mitigating, li2023evaluating,fu2023mme, zhao2024evaluating, thrush2022winoground}. 
{Our work improves the robustness of the VLMs to those issues by broadening visual capabilities of the VLMs.}

\section{Conclusions and Limitations}

In this paper, we focus on broadening the visual encoding aspect of VLMs. We first perform a comprehensive evaluation of several vision encoders on solving VLM tasks. Our findings revealed that there is no encoder achieving a consistent top performance across tasks and encoders with different biases can perform similarly. Motivated by this, we introduce \method that empowers VLMs by combining diverse features from several vision encoders using a small number of trainable parameters. Extensive evaluations on a broad range of tasks show that \method achieves state-of-the-art performance across the board and improves robustness of VLMs against out-of-distribution inputs and visual hallucinations. Below we discuss some limitations of our work and possible future directions:

\begin{itemize}
    \item \textit{Adaptive mechanisms}: While \bridge can resample from the useful features and discard the irrelevant ones, it still needs forward passes from all the encoders. A future direction could be to explore adaptive mechanisms to \textit{pre-select} which encoders to resample from, reducing the inference cost. 
    \item \textit{Improving sample efficiency}: \method uses around 100M image-text pairs during pre-training which is an order of magnitude less than recent methods~\cite{chen2022pali, wang2022git,bai2023qwen}. Further lowering the sample complexity is an important direction that will help reducing the development costs, e.g. by using a strategy similar to P-Former~\cite{jian2024bootstrapping}. 
    \item \textit{More broad vision encoders:} The set of encoders we investigated do not  completely cover all types of visual biases, e.g. inclusion of those with strong 3D~\cite{kar20223d, Eftekhar2021Omnidata} or semantic~\cite{kirillov2023segment} priors could further improve the performance. Our results show that the current set is a good starting point to see the benefits of consolidation. Another interesting direction is to strengthen vision encoders by distilling the \bridge-combined features back into a single encoder.  
    \item {\textit{Different modalities and multiple frames.} While we focus on image and text modalities, \method is a general method that could be potentially used to fuse more modalities, e.g. audio or 3D, or extended to incorporate multiple frames, e.g. for video understanding or in-context learning. 
    }
    \item \textit{Biases of text-generative models.} As we employ pre-trained LMs in our setup, the resulting VLMs are susceptible to the longstanding bias and fairness issues of these models, thus they should be used with caution especially for safety-critical applications.
    
\end{itemize}

\section{Acknowledgements}
We thank Diego Martin Arroyo, Ferjad Naeem, Xingyi Zhou, Yannick Strümpler and Yongqin Xian for their help with the project.

%
%
\bibliographystyle{splncs04}
\bibliography{main}

\begin{thebibliography}{100}
\providecommand{\url}[1]{\texttt{#1}}
\providecommand{\urlprefix}{URL }
\providecommand{\doi}[1]{https://doi.org/#1}

\bibitem{achiam2023gpt}
Achiam, J., Adler, S., Agarwal, S., Ahmad, L., Akkaya, I., Aleman, F.L., Almeida, D., Altenschmidt, J., Altman, S., Anadkat, S., et~al.: Gpt-4 technical report. arXiv preprint arXiv:2303.08774  (2023)

\bibitem{agrawal2019nocaps}
Agrawal, H., Desai, K., Wang, Y., Chen, X., Jain, R., Johnson, M., Batra, D., Parikh, D., Lee, S., Anderson, P.: Nocaps: Novel object captioning at scale. In: Proceedings of the IEEE/CVF international conference on computer vision. pp. 8948--8957 (2019)

\bibitem{Alayrac2022Flamingo}
Alayrac, J.B., Donahue, J., Luc, P., Miech, A., Barr, I., Hasson, Y., Lenc, K., Mensch, A., Millican, K., Reynolds, M., et~al.: Flamingo: a visual language model for few-shot learning. Advances in Neural Information Processing Systems  \textbf{35},  23716--23736 (2022)

\bibitem{awadalla2023openflamingo}
Awadalla, A., Gao, I., Gardner, J., Hessel, J., Hanafy, Y., Zhu, W., Marathe, K., Bitton, Y., Gadre, S., Sagawa, S., et~al.: Openflamingo: An open-source framework for training large autoregressive vision-language models. arXiv preprint arXiv:2308.01390  (2023)

\bibitem{bai2023qwen}
Bai, J., Bai, S., Chu, Y., Cui, Z., Dang, K., Deng, X., Fan, Y., Ge, W., Han, Y., Huang, F., et~al.: Qwen technical report. arXiv preprint arXiv:2309.16609  (2023)

\bibitem{bai2023sequential}
Bai, Y., Geng, X., Mangalam, K., Bar, A., Yuille, A., Darrell, T., Malik, J., Efros, A.A.: Sequential modeling enables scalable learning for large vision models. arXiv preprint arXiv:2312.00785  (2023)

\bibitem{bang2023multitask}
Bang, Y., Cahyawijaya, S., Lee, N., Dai, W., Su, D., Wilie, B., Lovenia, H., Ji, Z., Yu, T., Chung, W., et~al.: A multitask, multilingual, multimodal evaluation of chatgpt on reasoning, hallucination, and interactivity. arXiv preprint arXiv:2302.04023  (2023)

\bibitem{brock2021high}
Brock, A., De, S., Smith, S.L., Simonyan, K.: High-performance large-scale image recognition without normalization. In: International Conference on Machine Learning. pp. 1059--1071. PMLR (2021)

\bibitem{kakaobrain2022coyo-700m}
Byeon, M., Park, B., Kim, H., Lee, S., Baek, W., Kim, S.: Coyo-700m: Image-text pair dataset. \url{https://github.com/kakaobrain/coyo-dataset} (2022)

\bibitem{caron2021emerging}
Caron, M., Touvron, H., Misra, I., J{\'e}gou, H., Mairal, J., Bojanowski, P., Joulin, A.: Emerging properties in self-supervised vision transformers. In: Proceedings of the IEEE/CVF international conference on computer vision. pp. 9650--9660 (2021)

\bibitem{changpinyo2022all}
Changpinyo, S., Kukliansky, D., Szpektor, I., Chen, X., Ding, N., Soricut, R.: All you may need for vqa are image captions. arXiv preprint arXiv:2205.01883  (2022)

\bibitem{Changpinyo2021CC12M}
Changpinyo, S., Sharma, P.K., Ding, N., Soricut, R.: Conceptual 12m: Pushing web-scale image-text pre-training to recognize long-tail visual concepts. 2021 IEEE/CVF Conference on Computer Vision and Pattern Recognition (CVPR) pp. 3557--3567 (2021)

\bibitem{chen2023minigpt}
Chen, J., Zhu, D., Shen, X., Li, X., Liu, Z., Zhang, P., Krishnamoorthi, R., Chandra, V., Xiong, Y., Elhoseiny, M.: Minigpt-v2: large language model as a unified interface for vision-language multi-task learning. arXiv preprint arXiv:2310.09478  (2023)

\bibitem{chen2023shikra}
Chen, K., Zhang, Z., Zeng, W., Zhang, R., Zhu, F., Zhao, R.: Shikra: Unleashing multimodal llm's referential dialogue magic. arXiv preprint arXiv:2306.15195  (2023)

\bibitem{Chen2021Pix2seqAL}
Chen, T., Saxena, S., Li, L., Fleet, D.J., Hinton, G.: Pix2seq: A language modeling framework for object detection. In: International Conference on Learning Representations (2022)

\bibitem{chen2023pali}
Chen, X., Djolonga, J., Padlewski, P., Mustafa, B., Changpinyo, S., Wu, J., Ruiz, C.R., Goodman, S., Wang, X., Tay, Y., et~al.: Pali-x: On scaling up a multilingual vision and language model. arXiv preprint arXiv:2305.18565  (2023)

\bibitem{chen2022pali}
Chen, X., Wang, X., Changpinyo, S., Piergiovanni, A., Padlewski, P., Salz, D., Goodman, S., Grycner, A., Mustafa, B., Beyer, L., et~al.: Pali: A jointly-scaled multilingual language-image model. arXiv preprint arXiv:2209.06794  (2022)

\bibitem{chen2015microsoft}
Chen, X., Fang, H., Lin, T.Y., Vedantam, R., Gupta, S., Doll{\'a}r, P., Zitnick, C.L.: Microsoft coco captions: Data collection and evaluation server. arXiv preprint arXiv:1504.00325  (2015)

\bibitem{cherti2023reproducible}
Cherti, M., Beaumont, R., Wightman, R., Wortsman, M., Ilharco, G., Gordon, C., Schuhmann, C., Schmidt, L., Jitsev, J.: Reproducible scaling laws for contrastive language-image learning. In: Proceedings of the IEEE/CVF Conference on Computer Vision and Pattern Recognition. pp. 2818--2829 (2023)

\bibitem{chiang2023vicuna}
Chiang, W.L., Li, Z., Lin, Z., Sheng, Y., Wu, Z., Zhang, H., Zheng, L., Zhuang, S., Zhuang, Y., Gonzalez, J.E., et~al.: Vicuna: An open-source chatbot impressing gpt-4 with 90\%* chatgpt quality. See https://vicuna. lmsys. org (accessed 14 April 2023)  (2023)

\bibitem{cho2021unifying}
Cho, J., Lei, J., Tan, H., Bansal, M.: Unifying vision-and-language tasks via text generation. In: International Conference on Machine Learning. pp. 1931--1942. PMLR (2021)

\bibitem{chung2022scaling}
Chung, H.W., Hou, L., Longpre, S., Zoph, B., Tay, Y., Fedus, W., Li, Y., Wang, X., Dehghani, M., Brahma, S., et~al.: Scaling instruction-finetuned language models. arXiv preprint arXiv:2210.11416  (2022)

\bibitem{dai2023instructblip}
Dai, W., Li, J., Li, D., Tiong, A., Zhao, J., Wang, W., Li, B., Fung, P., Hoi, S.: Instruct{BLIP}: Towards general-purpose vision-language models with instruction tuning. In: Thirty-seventh Conference on Neural Information Processing Systems (2023), \url{https://openreview.net/forum?id=vvoWPYqZJA}

\bibitem{dehghani2022scenic}
Dehghani, M., Gritsenko, A., Arnab, A., Minderer, M., Tay, Y.: Scenic: A jax library for computer vision research and beyond. In: Proceedings of the IEEE/CVF Conference on Computer Vision and Pattern Recognition. pp. 21393--21398 (2022)

\bibitem{Dosovitskiy2020vit}
Dosovitskiy, A., Beyer, L., Kolesnikov, A., Weissenborn, D., Zhai, X., Unterthiner, T., Dehghani, M., Minderer, M., Heigold, G., Gelly, S., Uszkoreit, J., Houlsby, N.: An image is worth 16x16 words: Transformers for image recognition at scale. In: International Conference on Learning Representations (2021)

\bibitem{driess2023palm}
Driess, D., Xia, F., Sajjadi, M.S., Lynch, C., Chowdhery, A., Ichter, B., Wahid, A., Tompson, J., Vuong, Q., Yu, T., et~al.: Palm-e: An embodied multimodal language model. arXiv preprint arXiv:2303.03378  (2023)

\bibitem{Eftekhar2021Omnidata}
Eftekhar, A., Sax, A., Bachmann, R., Malik, J., Zamir, A.R.: Omnidata: A scalable pipeline for making multi-task mid-level vision datasets from 3d scans. 2021 IEEE/CVF International Conference on Computer Vision (ICCV) pp. 10766--10776 (2021)

\bibitem{fang2023eva}
Fang, Y., Wang, W., Xie, B., Sun, Q., Wu, L., Wang, X., Huang, T., Wang, X., Cao, Y.: Eva: Exploring the limits of masked visual representation learning at scale. In: Proceedings of the IEEE/CVF Conference on Computer Vision and Pattern Recognition. pp. 19358--19369 (2023)

\bibitem{Fang2022EVA}
Fang, Y., Wang, W., Xie, B., Sun, Q.S., Wu, L.Y., Wang, X., Huang, T., Wang, X., Cao, Y.: {EVA}: Exploring the limits of masked visual representation learning at scale. ArXiv  \textbf{abs/2211.07636} (2022)

\bibitem{fu2023mme}
Fu, C., Chen, P., Shen, Y., Qin, Y., Zhang, M., Lin, X., Yang, J., Zheng, X., Li, K., Sun, X., et~al.: Mme: A comprehensive evaluation benchmark for multimodal large language models. arXiv preprint arXiv:2306.13394  (2023)

\bibitem{geman1992neural}
Geman, S., Bienenstock, E., Doursat, R.: Neural networks and the bias/variance dilemma. Neural computation  \textbf{4}(1),  1--58 (1992)

\bibitem{gong2023multimodal}
Gong, T., Lyu, C., Zhang, S., Wang, Y., Zheng, M., Zhao, Q., Liu, K., Zhang, W., Luo, P., Chen, K.: Multimodal-gpt: A vision and language model for dialogue with humans. arXiv preprint arXiv:2305.04790  (2023)

\bibitem{goodwin1995seeing}
Goodwin, C.: Seeing in depth. Social studies of science  \textbf{25}(2),  237--274 (1995)

\bibitem{goyal2017making}
Goyal, Y., Khot, T., Summers-Stay, D., Batra, D., Parikh, D.: Making the v in vqa matter: Elevating the role of image understanding in visual question answering. In: Proceedings of the IEEE conference on computer vision and pattern recognition. pp. 6904--6913 (2017)

\bibitem{guo2023close}
Guo, B., Zhang, X., Wang, Z., Jiang, M., Nie, J., Ding, Y., Yue, J., Wu, Y.: How close is chatgpt to human experts? comparison corpus, evaluation, and detection. arXiv preprint arXiv:2301.07597  (2023)

\bibitem{guo2023images}
Guo, J., Li, J., Li, D., Tiong, A.M.H., Li, B., Tao, D., Hoi, S.: From images to textual prompts: Zero-shot visual question answering with frozen large language models. In: Proceedings of the IEEE/CVF Conference on Computer Vision and Pattern Recognition. pp. 10867--10877 (2023)

\bibitem{gurari2018vizwiz}
Gurari, D., Li, Q., Stangl, A.J., Guo, A., Lin, C., Grauman, K., Luo, J., Bigham, J.P.: Vizwiz grand challenge: Answering visual questions from blind people. In: Proceedings of the IEEE conference on computer vision and pattern recognition. pp. 3608--3617 (2018)

\bibitem{han2023imagebind}
Han, J., Zhang, R., Shao, W., Gao, P., Xu, P., Xiao, H., Zhang, K., Liu, C., Wen, S., Guo, Z., et~al.: Imagebind-llm: Multi-modality instruction tuning. arXiv preprint arXiv:2309.03905  (2023)

\bibitem{He2021MaskedAA}
He, K., Chen, X., Xie, S., Li, Y., Doll'ar, P., Girshick, R.B.: Masked autoencoders are scalable vision learners. 2022 IEEE/CVF Conference on Computer Vision and Pattern Recognition (CVPR) pp. 15979--15988 (2021)

\bibitem{heek1flax}
Heek, J., Levskaya, A., Oliver, A., Ritter, M., Rondepierre, B., Steiner, A., van Zee, M.: Flax: A neural network library and ecosystem for jax, 2020. URL http://github. com/google/flax  \textbf{1} (2020)

\bibitem{hoffmann2022training}
Hoffmann, J., Borgeaud, S., Mensch, A., Buchatskaya, E., Cai, T., Rutherford, E., Casas, D.d.L., Hendricks, L.A., Welbl, J., Clark, A., et~al.: Training compute-optimal large language models. arXiv preprint arXiv:2203.15556  (2022)

\bibitem{howard1995binocular}
Howard, I.P., Rogers, B.J.: Binocular vision and stereopsis. Oxford University Press, USA (1995)

\bibitem{Hu2021LoRA}
Hu, J.E., Shen, Y., Wallis, P., Allen-Zhu, Z., Li, Y., Wang, S., Chen, W.: {LoRA}: Low-rank adaptation of large language models. ArXiv  \textbf{abs/2106.09685} (2021), \url{https://api.semanticscholar.org/CorpusID:235458009}

\bibitem{huang2024visual}
Huang, W., Liu, H., Guo, M., Gong, N.Z.: Visual hallucinations of multi-modal large language models. arXiv preprint arXiv:2402.14683  (2024)

\bibitem{hudson2019gqa}
Hudson, D.A., Manning, C.D.: Gqa: A new dataset for real-world visual reasoning and compositional question answering. In: Proceedings of the IEEE/CVF conference on computer vision and pattern recognition. pp. 6700--6709 (2019)

\bibitem{ilharco_gabriel_2021_5143773}
Ilharco, G., Wortsman, M., Wightman, R., Gordon, C., Carlini, N., Taori, R., Dave, A., Shankar, V., Namkoong, H., Miller, J., Hajishirzi, H., Farhadi, A., Schmidt, L.: Openclip  (Jul 2021). \doi{10.5281/zenodo.5143773}, \url{https://doi.org/10.5281/zenodo.5143773}, if you use this software, please cite it as below.

\bibitem{jia2021scaling}
Jia, C., Yang, Y., Xia, Y., Chen, Y.T., Parekh, Z., Pham, H., Le, Q., Sung, Y.H., Li, Z., Duerig, T.: Scaling up visual and vision-language representation learning with noisy text supervision. In: International conference on machine learning. pp. 4904--4916. PMLR (2021)

\bibitem{jian2024bootstrapping}
Jian, Y., Gao, C., Vosoughi, S.: Bootstrapping vision-language learning with decoupled language pre-training. Advances in Neural Information Processing Systems  \textbf{36} (2024)

\bibitem{kar20223d}
Kar, O.F., Yeo, T., Atanov, A., Zamir, A.: 3d common corruptions and data augmentation. In: Proceedings of the IEEE/CVF Conference on Computer Vision and Pattern Recognition. pp. 18963--18974 (2022)

\bibitem{karpathy2015deep}
Karpathy, A., Fei-Fei, L.: Deep visual-semantic alignments for generating image descriptions. In: Proceedings of the IEEE conference on computer vision and pattern recognition. pp. 3128--3137 (2015)

\bibitem{kirillov2023segment}
Kirillov, A., Mintun, E., Ravi, N., Mao, H., Rolland, C., Gustafson, L., Xiao, T., Whitehead, S., Berg, A.C., Lo, W.Y., et~al.: Segment anything. arXiv preprint arXiv:2304.02643  (2023)

\bibitem{korbar2023text}
Korbar, B., Xian, Y., Tonioni, A., Zisserman, A., Tombari, F.: Text-conditioned resampler for long form video understanding. arXiv preprint arXiv:2312.11897  (2023)

\bibitem{li2023blip2}
Li, J., Li, D., Savarese, S., Hoi, S.: Blip-2: Bootstrapping language-image pre-training with frozen image encoders and large language models. arXiv preprint arXiv:2301.12597  (2023)

\bibitem{li2022blip}
Li, J., Li, D., Xiong, C., Hoi, S.C.H.: {BLIP}: Bootstrapping language-image pre-training for unified vision-language understanding and generation. In: International Conference on Machine Learning (2022)

\bibitem{li2021align}
Li, J., Selvaraju, R., Gotmare, A., Joty, S., Xiong, C., Hoi, S.C.H.: Align before fuse: Vision and language representation learning with momentum distillation. Advances in neural information processing systems  \textbf{34},  9694--9705 (2021)

\bibitem{li2023evaluating}
Li, Y., Du, Y., Zhou, K., Wang, J., Zhao, W.X., Wen, J.R.: Evaluating object hallucination in large vision-language models. arXiv preprint arXiv:2305.10355  (2023)

\bibitem{lin2023vila}
Lin, J., Yin, H., Ping, W., Lu, Y., Molchanov, P., Tao, A., Mao, H., Kautz, J., Shoeybi, M., Han, S.: Vila: On pre-training for visual language models. arXiv preprint arXiv:2312.07533  (2023)

\bibitem{lin2023sphinx}
Lin, Z., Liu, C., Zhang, R., Gao, P., Qiu, L., Xiao, H., Qiu, H., Lin, C., Shao, W., Chen, K., et~al.: Sphinx: The joint mixing of weights, tasks, and visual embeddings for multi-modal large language models. arXiv preprint arXiv:2311.07575  (2023)

\bibitem{liu2023mitigating}
Liu, F., Lin, K., Li, L., Wang, J., Yacoob, Y., Wang, L.: Mitigating hallucination in large multi-modal models via robust instruction tuning. In: The Twelfth International Conference on Learning Representations (2023)

\bibitem{liu2024survey}
Liu, H., Xue, W., Chen, Y., Chen, D., Zhao, X., Wang, K., Hou, L., Li, R., Peng, W.: A survey on hallucination in large vision-language models. arXiv preprint arXiv:2402.00253  (2024)

\bibitem{liu2023improved}
Liu, H., Li, C., Li, Y., Lee, Y.J.: Improved baselines with visual instruction tuning. arXiv preprint arXiv:2310.03744  (2023)

\bibitem{liu2023visual}
Liu, H., Li, C., Wu, Q., Lee, Y.J.: Visual instruction tuning. arXiv preprint arXiv:2304.08485  (2023)

\bibitem{Liu2023Prismer}
Liu, S., Fan, L.J., Johns, E., Yu, Z., Xiao, C., Anandkumar, A.: Prismer: A vision-language model with an ensemble of experts. ArXiv  \textbf{abs/2303.02506} (2023)

\bibitem{lu2023unified}
Lu, J., Clark, C., Lee, S., Zhang, Z., Khosla, S., Marten, R., Hoiem, D., Kembhavi, A.: Unified-io 2: Scaling autoregressive multimodal models with vision, language, audio, and action. arXiv preprint arXiv:2312.17172  (2023)

\bibitem{Lu2022UnifiedIOAU}
Lu, J., Clark, C., Zellers, R., Mottaghi, R., Kembhavi, A.: {Unified}-{IO}: A unified model for vision, language, and multi-modal tasks. In: The Eleventh International Conference on Learning Representations (2023)

\bibitem{marino2019ok}
Marino, K., Rastegari, M., Farhadi, A., Mottaghi, R.: Ok-vqa: A visual question answering benchmark requiring external knowledge. In: Proceedings of the IEEE/cvf conference on computer vision and pattern recognition. pp. 3195--3204 (2019)

\bibitem{4m}
Mizrahi, D., Bachmann, R., Kar, O.F., Yeo, T., Gao, M., Dehghan, A., Zamir, A.: {4M}: Massively multimodal masked modeling. In: Advances in Neural Information Processing Systems (2023)

\bibitem{naeem2023silc}
Naeem, M.F., Xian, Y., Zhai, X., Hoyer, L., Van~Gool, L., Tombari, F.: Silc: Improving vision language pretraining with self-distillation. arXiv preprint arXiv:2310.13355  (2023)

\bibitem{oquab2023dinov2}
Oquab, M., Darcet, T., Moutakanni, T., Vo, H.Q., Szafraniec, M., Khalidov, V., Fernandez, P., Haziza, D., Massa, F., El-Nouby, A., Assran, M., Ballas, N., Galuba, W., Howes, R., Huang, P.Y.B., Li, S.W., Misra, I., Rabbat, M.G., Sharma, V., Synnaeve, G., Xu, H., J{\'e}gou, H., Mairal, J., Labatut, P., Joulin, A., Bojanowski, P.: {DINOv2}: Learning robust visual features without supervision. ArXiv  \textbf{abs/2304.07193} (2023)

\bibitem{radford2021clip}
Radford, A., Kim, J.W., Hallacy, C., Ramesh, A., Goh, G., Agarwal, S., Sastry, G., Askell, A., Mishkin, P., Clark, J., et~al.: Learning transferable visual models from natural language supervision. In: International Conference on Machine Learning (2021)

\bibitem{Raffel2019T5}
Raffel, C., Shazeer, N.M., Roberts, A., Lee, K., Narang, S., Matena, M., Zhou, Y., Li, W., Liu, P.J.: Exploring the limits of transfer learning with a unified text-to-text transformer. The Journal of Machine Learning Research  \textbf{21}(1),  5485--5551 (2020)

\bibitem{Schuhmann2022LAION5B}
Schuhmann, C., Beaumont, R., Vencu, R., Gordon, C., Wightman, R., Cherti, M., Coombes, T., Katta, A., Mullis, C., Wortsman, M., Schramowski, P., Kundurthy, S., Crowson, K., Schmidt, L., Kaczmarczyk, R., Jitsev, J.: {LAION-5B}: An open large-scale dataset for training next generation image-text models. Advances in Neural Information Processing Systems  \textbf{35},  25278--25294 (2022)

\bibitem{schuhmann2022laion}
Schuhmann, C., Beaumont, R., Vencu, R., Gordon, C., Wightman, R., Cherti, M., Coombes, T., Katta, A., Mullis, C., Wortsman, M., et~al.: Laion-5b: An open large-scale dataset for training next generation image-text models. Advances in Neural Information Processing Systems  \textbf{35},  25278--25294 (2022)

\bibitem{shen2023chatgpt}
Shen, Y., Heacock, L., Elias, J., Hentel, K.D., Reig, B., Shih, G., Moy, L.: Chatgpt and other large language models are double-edged swords (2023)

\bibitem{team2023gemini}
Team, G., Anil, R., Borgeaud, S., Wu, Y., Alayrac, J.B., Yu, J., Soricut, R., Schalkwyk, J., Dai, A.M., Hauth, A., et~al.: Gemini: a family of highly capable multimodal models. arXiv preprint arXiv:2312.11805  (2023)

\bibitem{thorp2023chatgpt}
Thorp, H.H.: Chatgpt is fun, but not an author (2023)

\bibitem{thrush2022winoground}
Thrush, T., Jiang, R., Bartolo, M., Singh, A., Williams, A., Kiela, D., Ross, C.: Winoground: Probing vision and language models for visio-linguistic compositionality. In: Proceedings of the IEEE/CVF Conference on Computer Vision and Pattern Recognition. pp. 5238--5248 (2022)

\bibitem{tiong2022plug}
Tiong, A.M.H., Li, J., Li, B., Savarese, S., Hoi, S.C.: Plug-and-play vqa: Zero-shot vqa by conjoining large pretrained models with zero training. arXiv preprint arXiv:2210.08773  (2022)

\bibitem{tong2024eyes}
Tong, S., Liu, Z., Zhai, Y., Ma, Y., LeCun, Y., Xie, S.: Eyes wide shut? exploring the visual shortcomings of multimodal llms. arXiv preprint arXiv:2401.06209  (2024)

\bibitem{touvron2023llama}
Touvron, H., Lavril, T., Izacard, G., Martinet, X., Lachaux, M.A., Lacroix, T., Rozi{\`e}re, B., Goyal, N., Hambro, E., Azhar, F., et~al.: Llama: Open and efficient foundation language models. arXiv preprint arXiv:2302.13971  (2023)

\bibitem{vedantam2015cider}
Vedantam, R., Lawrence~Zitnick, C., Parikh, D.: Cider: Consensus-based image description evaluation. In: Proceedings of the IEEE conference on computer vision and pattern recognition. pp. 4566--4575 (2015)

\bibitem{wang2022git}
Wang, J., Yang, Z., Hu, X., Li, L., Lin, K., Gan, Z., Liu, Z., Liu, C., Wang, L.: Git: A generative image-to-text transformer for vision and language. arXiv preprint arXiv:2205.14100  (2022)

\bibitem{wang2022ofa}
Wang, P., Yang, A., Men, R., Lin, J., Bai, S., Li, Z., Ma, J., Zhou, C., Zhou, J., Yang, H.: Ofa: Unifying architectures, tasks, and modalities through a simple sequence-to-sequence learning framework. In: International Conference on Machine Learning. pp. 23318--23340. PMLR (2022)

\bibitem{wang2023cogvlm}
Wang, W., Lv, Q., Yu, W., Hong, W., Qi, J., Wang, Y., Ji, J., Yang, Z., Zhao, L., Song, X., et~al.: Cogvlm: Visual expert for pretrained language models. arXiv preprint arXiv:2311.03079  (2023)

\bibitem{Wang2022BEiT3}
Wang, W., Bao, H., Dong, L., Bjorck, J., Peng, Z., Liu, Q., Aggarwal, K., Mohammed, O.K., Singhal, S., Som, S., Wei, F.: Image as a foreign language: {BEiT} pretraining for all vision and vision-language tasks. ArXiv  \textbf{abs/2208.10442} (2022)

\bibitem{wang2021simvlm}
Wang, Z., Yu, J., Yu, A.W., Dai, Z., Tsvetkov, Y., Cao, Y.: Simvlm: Simple visual language model pretraining with weak supervision. arXiv preprint arXiv:2108.10904  (2021)

\bibitem{xu2023pixel}
Xu, J., Zhou, X., Yan, S., Gu, X., Arnab, A., Sun, C., Wang, X., Schmid, C.: {Pixel Aligned Language Models}. arXiv preprint arXiv: 2312.09237  (2023)

\bibitem{EvalAI}
Yadav, D., Jain, R., Agrawal, H., Chattopadhyay, P., Singh, T., Jain, A., Singh, S.B., Lee, S., Batra, D.: Evalai: Towards better evaluation systems for ai agents  (2019)

\bibitem{yao2021filip}
Yao, L., Huang, R., Hou, L., Lu, G., Niu, M., Xu, H., Liang, X., Li, Z., Jiang, X., Xu, C.: Filip: Fine-grained interactive language-image pre-training. arXiv preprint arXiv:2111.07783  (2021)

\bibitem{ye2023mplug}
Ye, Q., Xu, H., Xu, G., Ye, J., Yan, M., Zhou, Y., Wang, J., Hu, A., Shi, P., Shi, Y., et~al.: mplug-owl: Modularization empowers large language models with multimodality. arXiv preprint arXiv:2304.14178  (2023)

\bibitem{Yu2022CoCa}
Yu, J., Wang, Z., Vasudevan, V., Yeung, L., Seyedhosseini, M., Wu, Y.: Coca: Contrastive captioners are image-text foundation models. Transactions on Machine Learning Research  (2022)

\bibitem{yu2023self}
Yu, S., Cho, J., Yadav, P., Bansal, M.: Self-chained image-language model for video localization and question answering. arXiv preprint arXiv:2305.06988  (2023)

\bibitem{yuan2021florence}
Yuan, L., Chen, D., Chen, Y.L., Codella, N., Dai, X., Gao, J., Hu, H., Huang, X., Li, B., Li, C., et~al.: Florence: A new foundation model for computer vision. arXiv preprint arXiv:2111.11432  (2021)

\bibitem{zhai2022scaling}
Zhai, X., Kolesnikov, A., Houlsby, N., Beyer, L.: Scaling vision transformers. In: Proceedings of the IEEE/CVF Conference on Computer Vision and Pattern Recognition. pp. 12104--12113 (2022)

\bibitem{zhai2023sigmoid}
Zhai, X., Mustafa, B., Kolesnikov, A., Beyer, L.: Sigmoid loss for language image pre-training. arXiv preprint arXiv:2303.15343  (2023)

\bibitem{zhai2022lit}
Zhai, X., Wang, X., Mustafa, B., Steiner, A., Keysers, D., Kolesnikov, A., Beyer, L.: Lit: Zero-shot transfer with locked-image text tuning. In: Proceedings of the IEEE/CVF Conference on Computer Vision and Pattern Recognition. pp. 18123--18133 (2022)

\bibitem{zhang2024mm}
Zhang, D., Yu, Y., Li, C., Dong, J., Su, D., Chu, C., Yu, D.: Mm-llms: Recent advances in multimodal large language models. arXiv preprint arXiv:2401.13601  (2024)

\bibitem{zhang2022opt}
Zhang, S., Roller, S., Goyal, N., Artetxe, M., Chen, M., Chen, S., Dewan, C., Diab, M., Li, X., Lin, X.V., et~al.: Opt: Open pre-trained transformer language models. arXiv preprint arXiv:2205.01068  (2022)

\bibitem{zhao2024evaluating}
Zhao, Y., Pang, T., Du, C., Yang, X., Li, C., Cheung, N.M.M., Lin, M.: On evaluating adversarial robustness of large vision-language models. Advances in Neural Information Processing Systems  \textbf{36} (2024)

\bibitem{zhou2021ibot}
Zhou, J., Wei, C., Wang, H., Shen, W., Xie, C., Yuille, A., Kong, T.: {iBoT}: Image {BERT} pre-training with online tokenizer. In: International Conference on Learning Representations (2022)

\bibitem{zhu2023minigpt}
Zhu, D., Chen, J., Shen, X., Li, X., Elhoseiny, M.: Minigpt-4: Enhancing vision-language understanding with advanced large language models. arXiv preprint arXiv:2304.10592  (2023)

\end{thebibliography}
\end{document}